\documentclass[final]{cvpr}
\newcommand{\final}{0}

\usepackage{times}
\usepackage{epsfig}
\usepackage{graphicx}
\usepackage{bm}
\usepackage{amsmath}
\usepackage{amssymb}
\usepackage{amsthm}
\usepackage{subcaption}
\usepackage{url}
\usepackage{gensymb}
\usepackage[symbol]{footmisc}
\ifdefined\siggraph
\usepackage{times}
\fi

\usepackage{color}
\usepackage{ifthen}
\usepackage{float}
\usepackage{alltt}
\usepackage{newlfont} 
\usepackage{wrapfig}
\usepackage{booktabs}
\usepackage{multirow}
\usepackage{makecell}
\usepackage{array}
\usepackage{appendix}
\usepackage{comment}
\usepackage{amsmath}
\usepackage{amssymb}
\usepackage{amsthm}
\usepackage{textcomp}

\newcommand{\nothing}[1]{}

\definecolor{DeltaColor}{rgb}{0.039,0.73,0.71}
\definecolor{SetaColor}{rgb}{0.867, 0.0235, 0.376}
\definecolor{SigmaColor}{rgb}{0.98,0.45,0.0}
\definecolor{RedColor}{rgb}{0.8,0,0}
\definecolor{AlphaColor}{rgb}{0,0,0.8}
\definecolor{BetaColor}{rgb}{0.8,0,0.8}
\definecolor{GammaColor}{rgb}{0.5,0,0.7}
\definecolor{EpsilonColor}{rgb}{0.353,0.725,0.906}
\definecolor{TauColor}{rgb}{0.423,0.235,0.192}
\newcommand{\weikai}[1]{{\color{RedColor} Weikai: #1 $\qed$}}
\newcommand{\mingyue}[1]{{\color{AlphaColor} Mingyue: #1 $\qed$}}
\newcommand{\yuxin}[1]{{\color{SigmaColor} Yuxin: #1 $\qed$}}
\newcommand{\kui}[1]{{\color{GammaColor} Kui: #1 $\qed$}}

\newcommand{\warning}[1]{{\it\color{red} #1}}
\newcommand{\note}[1]{{\it\color{blue} #1}}

\definecolor{AudioColor}{rgb}{0.56,0.34,0.62}

\definecolor{DeadlineColor}{rgb}{0.9,0.4,0} 
\newcommand{\deadline}[1]{{\bf\color{DeadlineColor} ETA: #1}}

\definecolor{figred}{rgb}{1,0,0}
\definecolor{figgreen}{rgb}{0,0.6,0}
\definecolor{figblue}{rgb}{0,0,1}
\definecolor{figpink}{rgb}{1,0.63,0.63}

\newcolumntype{C}[1]{>{\centering}m{#1}}

\ifthenelse{\equal{\final}{1}}
{
\renewcommand{\weikai}[1]{}
\renewcommand{\mingyue}[1]{}
\renewcommand{\yuxin}[1]{}
\renewcommand{\kui}[1]{}
\renewcommand{\warning}[1]{}
\renewcommand{\note}[1]{}
\renewcommand{\deadline}[1]{}
}
{}

\newcounter{pccount}
\floatstyle{plain}

\newcommand{\filename}[1]{\url{#1}}
\newcommand{\foldername}[1]{\url{#1}}

\usepackage[pagebackref=true,breaklinks=true,colorlinks,bookmarks=false]{hyperref}

\def\cvprPaperID{6536} 
\def\confYear{CVPR 2021}

\graphicspath{
	{figs/raster/}
	{figs/handdrawn/}
}
 
\begin{document}
	%
	\title{Deep Optimized Priors for 3D Shape Modeling and Reconstruction}

 \author{Mingyue Yang$^1$, Yuxin Wen$^1$, Weikai Chen$^2$, Yongwei Chen$^1$,  Kui Jia$^1$\thanks{Corresponding author} \\  
$^1$South China University of Technology, $^2$Tencent America \\
{\tt\small \{eemingyueyang,wen.yuxin,eecyw\}@mail.scut.edu.cn},\\ {\tt\small chenwk891@gmail.com},
 {\tt\small kuijia@scut.edu.cn}
}    
	\maketitle
 
	\begin{abstract}
Many learning-based approaches have difficulty scaling to unseen data, as the generality of its learned prior is limited to the scale and variations of the training samples. This holds particularly true with 3D learning tasks, given the sparsity of 3D datasets available. We introduce a new learning framework for 3D modeling and reconstruction that greatly improves the generalization ability of a deep generator. Our approach strives to connect the good ends of both learning-based and optimization-based methods. In particular, unlike the common practice that fixes the pre-trained priors at test time, we propose to further optimize the learned prior and latent code according to the input physical measurements after the training. We show that the proposed strategy effectively breaks the barriers constrained by the pre-trained priors and could lead to high-quality adaptation to unseen data. We realize our framework using the implicit surface representation and validate the efficacy of our approach in a variety of challenging tasks that take highly sparse or collapsed observations as input. Experimental results show that our approach compares favorably with the state-of-the-art methods in terms of both generality and accuracy. 
\end{abstract}

	 \begin{figure}[t]
    \centering
 \includegraphics[width=.49\textwidth]{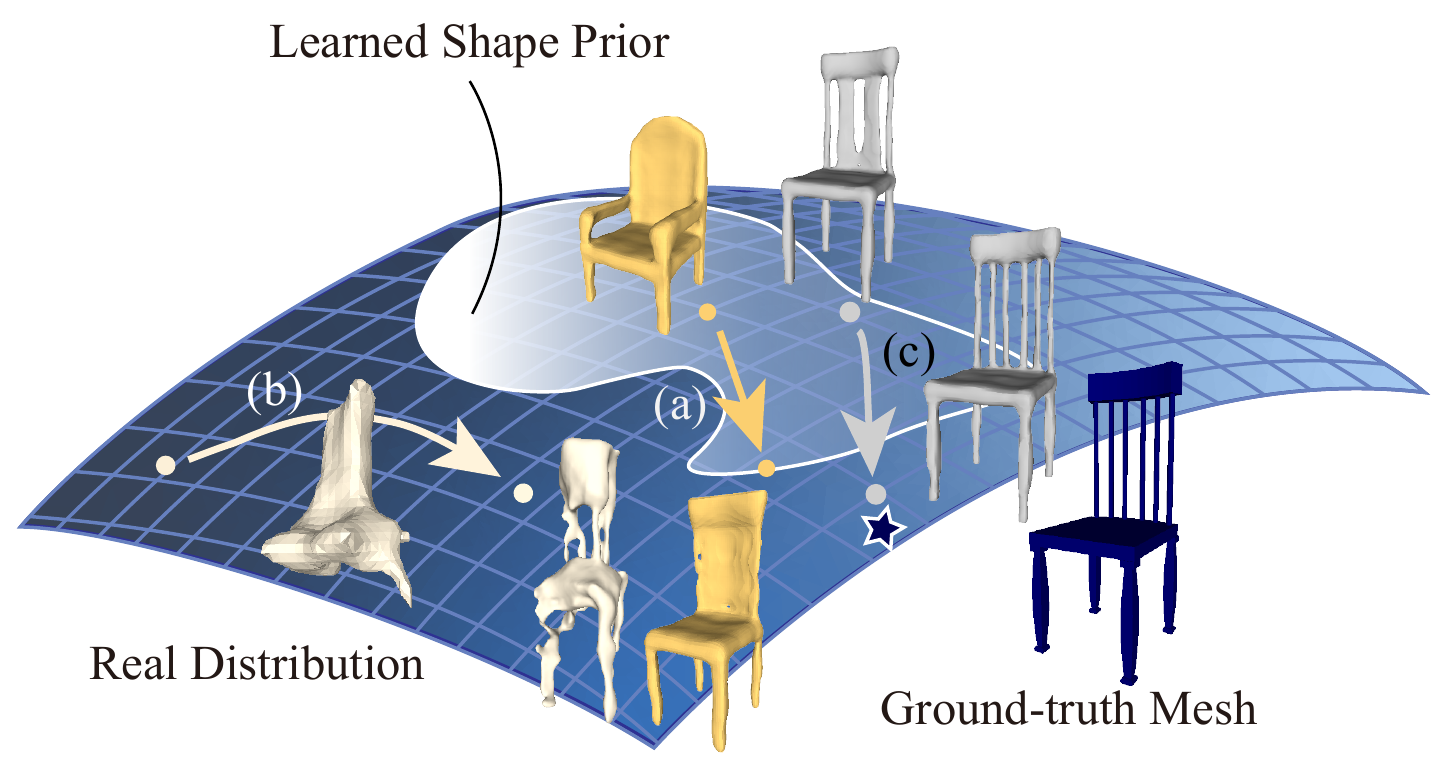}
    \caption{
    The shape prior learned from the limited training data cannot capture the full landscape of the real data distribution. 
    Common practice that uses a fixed pre-trained generator is constrained within the prior (path a) and thus fails to model the unseen data lying outside the prior, even with latent code optimization at test time. Optimizing a randomly initialized generator, on the other hand, is prone to be trapped in a local minimum due to the complex energy landscape (path b). 
    Whereas the pre-train prior could provide a good initialization in a forward pass, we propose to further optimize the parameters of the prior and the latent code according to the task-specific constraints at test time. We show in this work that the proposed framework can effectively break the barriers of pre-trained prior and generalize to the unseen data that is out of the prior domain (path c). 
    Hence, our approach can generate results (ending point of path c) closest to the ground truth (star point on the real data manifold) compared to the other learning methods (path a and b).
  }
    \label{fig:teaser}
    \vspace{-0.5cm}
\end{figure}
	\section{Introduction}
\label{sec:intro}

Deep generative models have brought impressive advances to the state-of-the-art across a wide variety of generative tasks, including 2D image synthesis and 3D shape reconstruction.
At the moment, it is widely believed that these leaps in performance come primarily from the realistic priors learned from a large amount of training data.
Based on this observation, most of the previous 3D learning approaches focus on learning stronger priors during training and strictly respecting the learned prior at test time.
Specifically, there are two common ways to leverage the learned shape priors. 
One is to train an encoder to retrieve the most likely prior by mapping the input into the latent code, which is a low-dimensional representation of the shape prior. 
The other is to optimize the latent code until its decoded output achieves a minimal loss.
Note that both methods fix the learned prior/generator once the training is completed as the prior is considered to be the most valuable asset in a learning-based approach.
However, is this the best strategy of using the prior in a 3D learning task?

The quality of the learned prior highly relies on the scale and diversities of the training examples. Yet, even with a large amount of data, the prior learned by the neural network may still be a crude approximation of the real data distribution (see Figure~\ref{fig:teaser}), making the network vulnerable to unseen data.
This is particularly true with the 3D learning tasks, where the ground-truths are notoriously difficult to obtain, which greatly limits the scale of the training samples.
Optimization-based approaches that leverage physically correct observations, e.g. multi-view consistency, do not require any training to be usable. 
However, they are strict with the inputs and tend to fail on the sparsity of the data (e.g. single/sparse-view reconstruction) or the physical misalignment (e.g. unregistered/mismatched images).

To alleviate the generality issue of the learning-based approach while maintaining a friendly requirement for the inputs, we advocate a new 3D learning paradigm that connects the good ends of both learning-based and optimization-based approaches. 
In particular, we propose that the pre-trained data prior could obtain a maximum generality if it is \textit{optimized}, rather than \textit{fixed}, according to the physical constraints at test time.
Our approach shares a similar incentive with deep image priors~\cite{ulyanov2018deep}, where high-quality images can be synthesized simply by optimizing an untrained and randomly initialized deep generator. 
However, unlike image synthesis, we show that optimizing a randomly initialized neural network often fails to achieve satisfactory results in 3D learning, especially in highly ill-posed configurations, such as sparse-view based 3D reconstruction.
 
Instead of fixing the priors or using random priors, we propose to jointly optimize the pre-trained shape prior and the latent code towards the input physical measurements at test time.
Our observation is that though the learned prior cannot capture the full landscape of the real data distribution, it does provide a fairly good initialization for searching for the optimal solution in the entire embedding space (Figure~\ref{fig:teaser}).
Further, by introducing the physically based optimization, the searching path could break the barrier of the pre-trained priors and converge at some point on the real prior which is more realistic but unreachable by only searching inside the learned priors (Figure~\ref{fig:teaser}).
While it is possible that the optimization may lead to 3D shapes that do not look plausible, we propose that an $l_2$ regularization works surprisingly well in regularizing the searching space.

We materialize our idea using the implicit surface representation, as it is flexible to handle shapes with arbitrary topologies.
We show that our proposed approach is a general 3D learning framework that supports a wide range of downstream applications, including shape modeling and reconstruction, with various forms of inputs.
We also demonstrate that our framework can significantly improve the generality of the learning-based approach, even in the presence of highly sparse or collapsed observations, e.g. the sparse point clouds obtained from the 3D scanning, single or sparse views of the object of interest, etc.
We verify the effectiveness of our approach in a variety of challenging tasks, including shape auto-encoding, sparse-view reconstruction and sparse point cloud reconstruction.
Experimental results show that our approach is superior to the state-of-the-arts both quantitatively and qualitatively. 
 \vspace{-0.1cm}
 
	\section{Related Work}
\label{sec:related_work}

\noindent\textbf{Optimization-based Shape Reconstruction.}
Traditional image-based surface reconstruction methods, including PMVS \cite{furukawa2007accurate} and COLMAP \cite{schonberger2016pixelwise}, \etc, are mainly based on texture-rich and dense views for extracting multi-view correspondences. 
Since these approaches follow the exact physical constraints, the reconstructed surface could be highly accurate.
Nonetheless, they are also vulnerable to noisy input and collapsed observations which could interrupt the acquisition of pixel-wise correspondence across different views.
In addition, they fail to generate plausible results in the presence of sparse views.
 The other line of research strives to reconstruct 3D surface from raw point clouds.
The most representative ones include Poisson surface reconstruction \cite{kazhdan2013screened}, radius basis functions (RBF) \cite{carr2001reconstruction}, and moving least squares (MLS) \cite{levin2004mesh} based approaches. 
The main idea of these methods is to fit either polygonal meshes or implicit functions to the input point cloud by optimizing a pre-defined energy objective.
In contrast, our proposed method take advantage of both learning-based and optimization-based framework.
Specifically, while we are able to faithfully reconstruct 3D surface with sparse or even a single view, we can also achieve similar quality of reconstruction with the traditional stereo-based approach when dense views are available. 
 \vspace{-0.15cm}
 \begin{figure*}[t]
    \centering
    \includegraphics[width=.90\textwidth]{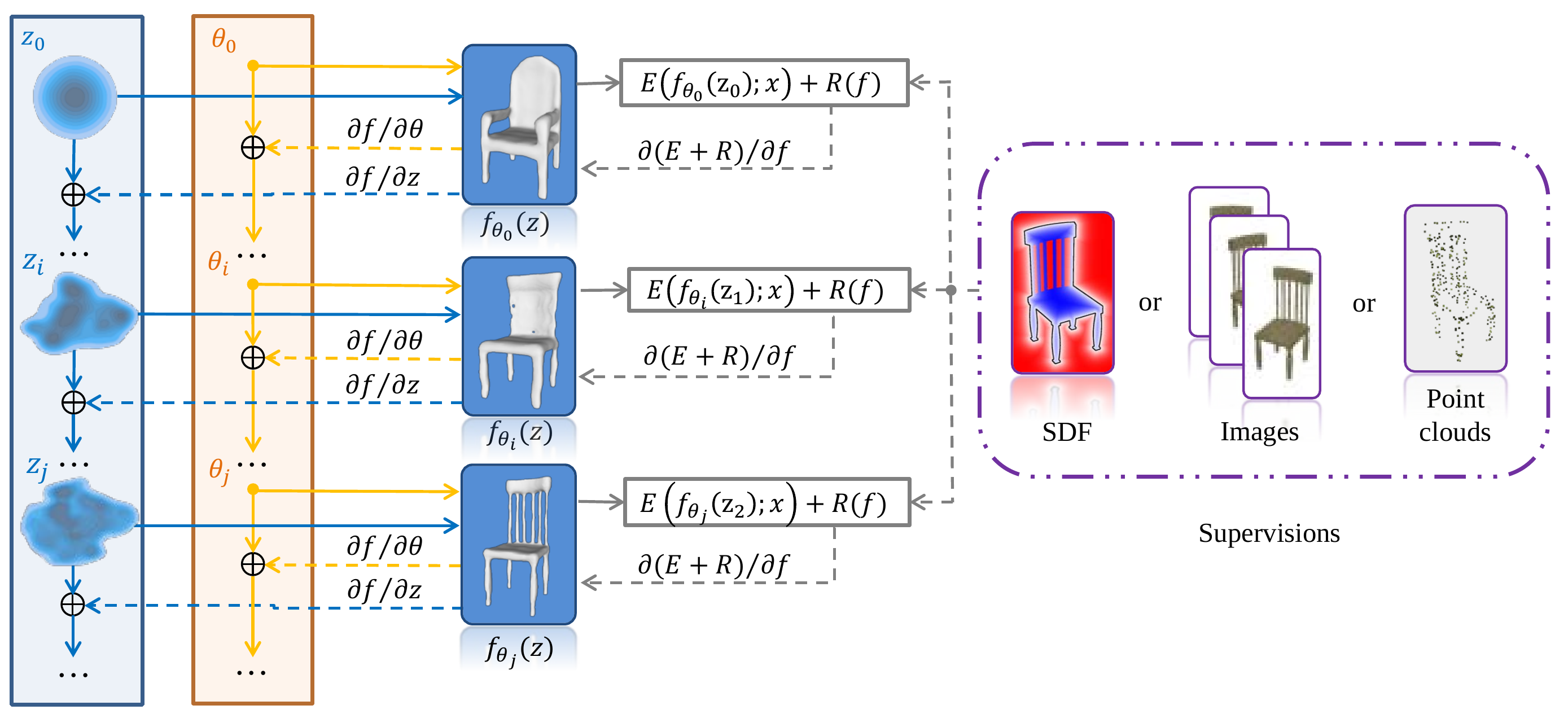}
    \vspace{-0.05cm}
    \caption{
    Illustration of our pipeline.
    Given specific supervisions (right) from the downstream applications, our goal is to optimize both the latent code $z$ and the parameters $\theta$ of the pre-trained shape prior to generate a high-fidelity 3D result (left), which is represented by a neural implicit field.
    We perform iterative optimizations that alternatively optimize $z$ and $\theta$ in each iteration.
    Our optimization framework can gradually break the barrier of the pre-trained shape prior and converge to a faithful 3D shape that could lie out of the manifold spanned by training samples.
    We visualize the distribution of latent code in the leftmost using t-SNE technique \cite{maaten2008visualizing} and the reconstructed results using marching cube \cite{lorensen1987marching} for better visualization.
    }
    \label{fig:framework}
    \vspace{-0.15cm}
\end{figure*}
\vspace{0.1cm}

\noindent\textbf{Learning-based Shape Modeling and Reconstruction.}
Recent years have witnessed great progress in introducing deep learning to 3D shape modeling and reconstruction.
In particular, most of the previous works mainly rely on a retrieval based framework that fixes the parameters of the generator after training and retrieves the closest prior in the latent space via forward passing.
It has been widely used in a wide range of 3D representations, including mesh~\cite{wang2018pixel2mesh,wen2019pixel2meshplus,groueix2018papier}, voxels~\cite{choy20163d,tatarchenko2017octree,learning-lsm}, and implicit field~\cite{mescheder2019occupancy,chen2019learning,NIPS2019_8340}.
Though reasonable results can be obtained from these methods, they are vulnerable to disturbances in the input. 
Once the forward pass failed, one cannot further modify or optimize the results.
To resolve this issue, recent works~\cite{achlioptas2018learning,liu2020dist} have proposed to optimize the latent code at test time.
DeepSDF~\cite{park2019deepsdf} presents the framework of auto-decoder where the shape prior is learned only with a decoder during training.
The latent codes are optimized according to the input observations at test time, given a fixed pre-trained decoder.
Despite that the reconstruction accuracy has been further improved by these approaches, they still have difficulty generalizing to unseen data as the pre-trained prior is limited to the domain spanned by the training samples.
Recently, Williams et al.~\cite{williams2019deep} propose to overfit a randomly initialized the neural network to an input point cloud. 
While surprisingly well results can be achieved in this setting, we show that it can hardly be applied to other challenging tasks, such as sparse-view surface reconstruction.
In this paper, we propose a more general learning framework that strives to optimize both the pre-trained shape prior and latent code at test time. We show that it can significantly improve the generality and performance of the deep learned model in a wide range of highly ill-posed problems. 
\vspace{0.1cm}

\noindent\textbf{Combination of learning with physical constraints.}
There have been a few preliminary explorations attempting to combine deep learning with optimization based on physical constraints.
In particular, \cite{zhu2018object} and \cite{lin2019photometric} strive to reduce the searching space of a traditional optimization problem using a deep learned prior.
They first encode the shape into latent space, and then optimize the latent code according to the photometric consistency constraint~\cite{lin2019photometric} or the bundle adjustment loss \cite{zhu2018object}.
Though results that are more physically plausible can be achieved with these approaches, their performance is still limited to the quality of pre-trained prior and thus struggles to scale to unseen data.
Another line of research aims to introduce physical constraints as the supervision signal during training.
For instance, the latest advances in differentiable rendering~\cite{liu2019softras,kato2018neural} have been widely used in achieving unsupervised learning of single-view 3D mesh reconstruction.
To leverage the flexibility of implicit representation, recent works~\cite{liu2019learning, DVR,liu2020dist,yariv2020multiview} have proposed new techniques to render implicit surfaces differentiably.
These approaches succeed in training more powerful priors as there are ample resources of 2D images that can be directly used for training.
However, they still do not resolve the generalization problem. 
Besides, since differentiable rendering techniques make 2D supervision possible, these methods ~\cite{DVR,yariv2020multiview} can generate a single shape using dense views without learnt prior. In these cases, they can be considered as optimization-based methods.
In contrast, our approach aims to incorporate the task-specific physical constraints for optimizing shape priors and latent codes at test time, which can further boost the performance and generality of the previous approaches.
\vspace{-0.1cm}
	\vspace{-0.15cm}
\section{Methods}
\label{sec:method}
We interpret the implicit function learned by a deep generator as $f_{\bm{\theta}}(\bm{x},\bm{z})$, which maps a query 3D point $\bm{x}$ and a latent code $\bm{z}$ to the target shape's approximate signed distance field.  Our goal is to generate or reconstruct a faithful 3D surface $\mathcal{O}$ from the input physical observations, \eg  spatial signed distance field $\mathcal{S}$, sparse multi-view images $\mathcal{I}$, or point clouds $\mathcal{P}$, \etc, by leveraging the priors encoded in $f_{\bm{\theta}}(\bm{x},\bm{z})$.
Given the decoding model $f_{\bm{\theta}}$, the continuous surface associated with a latent code $\bm{z}$ is represented by the decision boundary of $f_{\bm{\theta}}(\bm{x},\bm{z})$, and the shape can be instantiated by the marching cubes algorithm. 
Formally, a general 3D modeling problem can be formulated as follows:
\begin{equation}
    \hat{\mathcal{O}}^{*} = \arg\min_{\hat{\mathcal{O}}}E(\hat{\mathcal{O}}; \mathcal{O}) + R(\hat{\mathcal{O}}) , 
\end{equation}
\noindent where $E(\hat{\mathcal{O}}, \mathcal{O})$ is a task-specific energy term, and $R(\hat{\mathcal{O}})$ is a regularizer that encourages a plausible surface.
 
In deep image prior~\cite{ulyanov2018deep}, the regularizer $R(\hat{\mathcal{O}})$ is realized using a randomly initialized and untrained neural network.
However, unlike image synthesis, the 3D inverse problems are much harder, where merely depending on the prior brought by the structure of neural network is insufficient.
Therefore, instead of random initialization, we leverage a pre-training to initialize $\bm{\theta}$ and $\bm{z}$ more appropriately.
Hence, our goal is defined in the following form:
\begin{equation}
    \label{eq:overall_objective}
    \begin{split}
        \bm{z}^{*}, \bm{\theta}^{*} = &\arg\min_{\bm{z},\bm{\theta}} E(f_{\bm{\theta}}(\bm{x},\bm{z}); \mathcal{O}) + R(f_{\bm{\theta}}(\bm{x},\bm{z})),
    \end{split}
\end{equation}
\noindent where, instead of optimizing from scratch or a randomized neural network, we advocate to iteratively optimize the pre-trained priors, including the generator parameters $\bm{\theta}$ and the latent code $\bm{z}$, according to the physical constraints $E(f_{\bm{\theta}}(\bm{x},\bm{z}); \mathcal{O})$ (Figure~\ref{fig:framework}). Further, the learned prior can be used as a strong regularizer $R(f_{\bm{\theta}}(\bm{x},\bm{z}))$ to ensure a reasonable output.
The formulation in Eq.~(\ref{eq:overall_objective}) shows our general framework that combines the learning-based and optimization-based approaches.
We will show in the next section how this formulation can be adapted to various applications.
 
	\section{Applications}
\label{sec:applications}

We now show experimentally how the proposed approach works for diverse tasks on 3D modeling and reconstruction that take different input forms. 
Note that each application requires a pre-trained shape prior. Since the main focus of this work is not about how to obtain a stronger prior, we provide the details of our pre-training in the appendix.
\subsection{Shape Auto-Encoding}
\label{sec:sdfR}
Auto-encoding 3D shapes play an important role in obtaining shape priors and a variety of downstream applications related to shape modeling and reconstruction.
Since we implement our framework using implicit surface representation, our goal is to generate an implicit field  as a faithful approximation of the input surface $\mathcal{S}$.
We first convert the 3D locations to be queried into a signed distance field.
The resulted field is composed of a set of pair $\{(\bm{p}_i,s_i)\}_{i=1}^n$, where the first element is the coordinates of the querying position in the space and the second element is its corresponding distance value.
In particular, the reconstruction energy term in Eq. (\ref{eq:overall_objective}) is represented as 
\begin{equation}\label{eq:energy_sdf}
    E(f_{\bm{\theta}}(\bm{z}); \mathcal{X}) = \sum_{i\in\{1,\ldots,n\}} \|\hat{s}_i - s_i\|_1 ,
\end{equation}
\noindent where $s_i$ is the ground-truth distance; $\hat{\bm{s}}_{i}$ denotes the estimated signed distance value for the $i^{th}$ point $\bm{p}_i$, predicted via the neural implicit field $f_{\bm{\theta}}(\bm{z},\bm{p}_i)$.
We apply the regularizer as that in Eq. (\ref{eq:regu_mvs}), which will be discussed later in Section 4.2, namely  $R(f_{\bm{\theta}}(\bm{z}))$, to ensure high-fidelity and reasonable results.
This leads to a similar overall objective function as shown in Eq. (\ref{eq:overall_mvs}).
 
\subsection{Multi-view Reconstruction}
\label{sec:mvs}
Given a collection of multi-view images $\mathcal{I}$, together with object silhouette masks $\mathcal{M}$, the camera extrinsics $\mathcal{P}$ and intrinsics $\mathcal{K}$, the aim of multi-view reconstruction is to recover the underlying object surface from these partial observations of $n$ views.
 To correlate the 3D surface with the 2D observations, we leverage the differentiable rendering technique such that the renderings of the generated surface are consistent with the input views.
 For more details on the differentiable rendering technique we used, please refer to the appendix.
Thereby, the energy term in Eq~(\ref{eq:overall_objective}) is formulated as:
\begin{equation}\label{eq:energy_mvs}
    E(f_{\bm{\theta}}(\bm{z}); \mathcal{X}) = \sum_{i=1}^n (\|\hat{\bm{I}}_{i} - \bm{I}_{i} \|_1 + \lambda_c\cdot\mathcal{L}_{c}(\hat{\bm{M}}_{i} - \bm{M}_{i}))
\end{equation}
\noindent where $\mathcal{L}_{c}$ is the binary cross entropy, and $\lambda_c$ is the weighted parameter.
$\hat{\bm{I}}_{i}$ and $\hat{\bm{M}}_{i}$ denotes the estimated image and silhouette respectively for the $i^{th}$ view.
 Specifically, the first term restrains only on the pixels inside the intersection of the given mask $\bm{M}_i$ and the predicted mask $\hat{\bm{M}}_{i}$, where the photometric RGB loss can be defined reasonably; while the second term applies to all the pixels to penalize mismatched object silhouettes.

In the presence of highly sparse views, the multi-view reconstruction task becomes a highly underdertermined problem. 
Hence, we further introduce additional regularizers on the neural network to ensure plausible results.
For $\bm{z}$, we encourage the prior distribution of the latent code to be a zero-mean multivariate-Gaussian to encapsulate them into a compact shape manifold, preventing biased solutions.
 In addition, we would like to prevent $\bm{\theta}$ from moving too far away from the learned categorical prior.
Through extensive experiments, we find that a simple $l_2$ norm on $\bm{\theta}$ works surprisingly well to strike a balance between flexibility and regularity.
Formally, the regularizer term in Eq. (\ref{eq:overall_objective}) is defined as:
\begin{equation}\label{eq:regu_mvs}
    R(f_{\bm{\theta}}(\bm{z})) = \frac{1}{\sigma^2}\|\bm{z}\|_2 + \lambda_\theta\cdot\|\bm{\theta}-\bm{\theta}_0\|_2,
\end{equation}
where $\lambda_\theta$ denotes the weighted parameter, and $\bm{\theta}_0$ denotes the parameters of $\bm{\theta}$ learned from the pre-training dataset.
All together, the energy objective is formulated as:
\begin{equation}\label{eq:overall_mvs}
    \min_{\bm{\theta},\bm{z}} L(\mathcal{X})=E(f_{\bm{\theta}}(\bm{z}); \mathcal{X}) + \lambda\cdot R(f_{\bm{\theta}}(\bm{z})),
\end{equation}
where $\lambda$ is the regularizer parameter.
The overall default values are set as $\lambda=0.5$, $\lambda_c=0.5$, $\lambda_\theta=0.1$, which works well in all our experiments.

\subsection{Point Cloud Reconstruction}
\label{sec:pcR}
Our approach also supports reconstructing a complete 3D shape from the sparse 3D observation -- point cloud $\mathcal{P}$.
In this case, the input $\mathcal{X}$ is composed of a set of 3D points $\{\bm{p}_i\}_{i=1}^n$  with or without their corresponding normals $\{\bm{n}_i\}_{i=1}^n$.
The goal is to reconstruct the continuous implicit field to represent the plausible object surface $\mathcal{O}$ that best fit the inputs.
We hence can reformulate the energy term in Eq. (\ref{eq:overall_objective}) as:
\begin{equation}\label{eq:energy_pc}
    E(f_{\bm{\theta}}(\bm{z}); \mathcal{X}) = \sum_{i\in\{1,\ldots,n\}} (\|\hat{s}_i\|_1 + \lambda_n\cdot\|\hat{\bm{n}}_i-\bm{n}_i\|_2),
\end{equation}
\noindent where $\lambda_n$ is the weight for the normal regularizer;
$\hat{s}_i = f_{\bm{\theta}}(\bm{z},\bm{p}_i)$ and $\hat{\bm{n}}_i = \nabla_{\bm{p}}f_{\bm{\theta}}(\bm{z},\bm{p}_i)$
are the estimated signed distance value and normal for the $i^{th}$ point $\bm{p}_i$ respectively.
Note that the normal term is optional depending on the availability of the normal data.
 
To encourage a smooth surface, apart from the multivariate-Gaussian prior for the latent space, we also include an Eikonal term \cite{crandall1983viscosity}, which regularizes the $l_2$-norm of the gradients $\nabla_{\bm{p}}f(\bm{z},\bm{p}_i)$.
The regularization term can be formulated as: 
\begin{equation}\label{eq:regu_pc}
\begin{split}
    R(f_{\bm{\theta}}(\bm{z})) = \frac{1}{\sigma^2}&\|\bm{z}\|_2 + \lambda_\theta\cdot\|\bm{\theta}-\bm{\theta}_0\|_2 \\
    & + \lambda_g\cdot\mathbb{E}_{\bm{p}}(\|\nabla_{\bm{p}}f_{\bm{\theta}}(\bm{z},\bm{p}_i)\|_2-1)^2,
\end{split}
\end{equation}
\noindent where $\lambda_\theta$ and $\lambda_g$ are the weights for their regularization terms.
The Eikonal term is formulated as the expectation with respect to the probability distribution of $\bm{p}$.
As it encourages the gradients $\nabla_{\bm{p}}f_{\theta}$ to be of unit-2 norm, $f_\theta$ will achieve minimum loss of Eq. (\ref{eq:regu_pc}) if $f_\theta$ vanishes on $\bm{p}$ and becomes a signed distance in Euclidean metric.

	\section{Experimental Results}
\label{sec:experiments}

\vspace{0.1cm}
\noindent\textbf{Dataset.}
We adopt the category of chairs, lamps and cars in ShapeNet Core dataset (v2) \cite{chang2015shapenet} as our dataset, with $6778,2318,7497$ shapes respectively.
Each mesh is normalized into a unit sphere during pre-processing.
For the task of auto-encoding given input shape, we follow Park \etal \cite{park2019deepsdf} to construct the signed distance fields, each with $250,000$ spatial points and their values.
For surface reconstruction based on sparse input RGB images, we use the rendered dataset from the Choy \etal \cite{choy20163d} to adhere to the community standards \cite{DVR, mescheder2019occupancy, wang2018pixel2mesh, wen2019pixel2meshplus}. 
The dataset contains $24$ images of resolution $64^2$ and the viewpoints are sampled on the northern hemisphere of the object.
We use $24$ images and corresponding object masks per object for supervision on the pre-training stage to obtain a good prior, and $3$ images of the same resolution for testing.
As for the task of point cloud based reconstruction, we sample $150,000$ points and their corresponding normals for training shape priors, but only use $300$ points for evaluating our performance on sparse point cloud reconstruction.

\vspace{0.1cm}
\noindent\textbf{Evaluation metrics.}
For quantitative evaluations, we apply the most commonly used metrics of Chamfer Distance (CD) between uniformly sampled point clouds to measure the accuracy and completeness of the surface (the lower the better).
For shape auto-encoding, we adopt the median of Chamfer Distance (the lower the better) following \cite{park2019deepsdf}.
For multi-view reconstruction, following \cite{wen2019pixel2meshplus}, we further adopt F-Score, measuring the completeness and precision of generated shapes (the higher the better).
For point cloud reconstruction, we use normal consistency~\cite{chibane2020implicit} to measure the accuracy and completeness of the shape normals (the higher the better).
 \subsection{Shape Auto-Encoding}
\label{sec:res_ae}

\begin{figure}
    \centering
   
    \begin{minipage}[c]{0.08\textwidth}
        \raggedright
        DeepSDF  
    \end{minipage}
    \begin{minipage}[c]{0.38\textwidth}
        \raggedright 
        \includegraphics[height=0.31\textwidth]{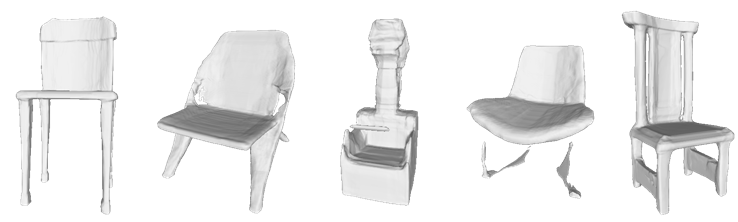}
    \end{minipage}\\
    \begin{minipage}[c]{0.08\textwidth}
        \raggedright
        Ours
    \end{minipage}
    \begin{minipage}[c]{0.38\textwidth}
        \raggedright 
        \includegraphics[height=0.31\textwidth]{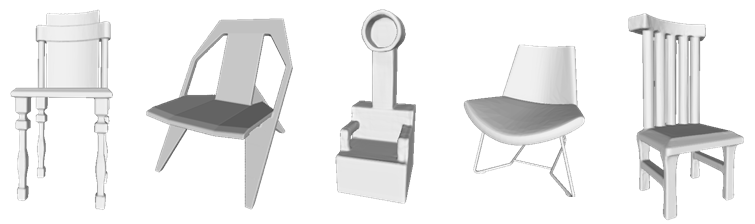}
    \end{minipage}\\
    \begin{minipage}[c]{0.08\textwidth}
        \raggedright
        GT
    \end{minipage}
    \begin{minipage}[c]{0.38\textwidth}
        \raggedright 
        \includegraphics[height=0.31\textwidth]{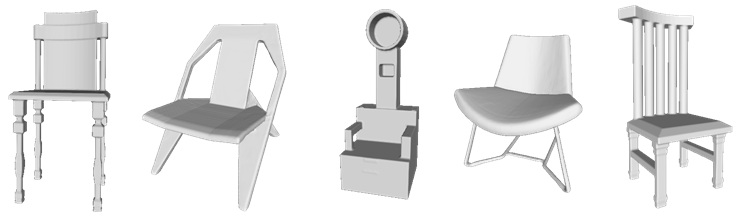}
    \end{minipage}
 
    \caption{
    Qualitative comparisons between shape auto-encoding results generated by different methods.}
    \label{fig:sdf_comparing}
    \vspace{-0.3cm}
\end{figure}

We compare the performance of shape auto-encoding with DeepSDF~\cite{park2019deepsdf} in this section.
We show the quantitative and qualitative results in Table~\ref{table:sdf_comparing} and Figure~\ref{fig:sdf_comparing} respectively.
 As shown in Figure \ref{fig:sdf_comparing}, our approach performs significantly better in recovering the fine details, such as the bumping details in the chair legs (1st column) and the thin rods on the chair back (5-th column).
This performance leap become more prominent when the testing object deviates stronger from the training set.
The quantitative results in Table \ref{table:sdf_comparing} further support that our method performs much better across a range of different instances compared to DeepSDF \cite{park2019deepsdf}.

\begin{table}
\begin{center}
\begin{tabular}{|l|c|c|c|c|}
\hline
\multirow{2}{*}{Methods} & \multicolumn{2}{c|}{CD, mean $(10^{-3})$} & \multicolumn{2}{c|}{CD, median $(10^{-3})$}  \\
\cline{2-3}\cline{4-5}
~ & Chair & Table & Chair & Table \\
\hline\hline
DeepSDF   &0.21 & 0.42 & 0.08& 0.07\\
\hline
Ours & $\bm{0.08}$ & $\bm{0.10}$ & $\bm{0.06}$ & $\bm{0.05}$\\
\hline
\end{tabular}
\end{center}
\vspace{-0.3cm}
\caption{
Comparative results of different methods for shape auto-encoding reconstruction.
Performance is measured in terms the mean value and median value of Chamfer distance over $50$ instances.
$10^{-3}$ refers to the magnitude.
}
\label{table:sdf_comparing}
\vspace{-0.4cm}
\end{table}

\subsection{Sparse Multi-view Reconstruction}
\label{sec:res_multiview}

 \begin{figure*}
    \begin{tabular}{
    p{50pt}p{50pt}p{50pt}p{50pt}p{60pt}p{50pt}p{50pt}p{50pt}
    }
    \ \ \ Images & \ \ \ LSM\cite{learning-lsm} & \  P2M++\cite{wen2019pixel2meshplus} & \ \  DISN\cite{NIPS2019_8340} & \ \ \ \ DVR\cite{DVR} &  IDR\cite{yariv2020multiview} &  Ours &GT 
    \end{tabular}\vspace{-0.1cm}\\
    \centering
   
    \includegraphics[width=.99\textwidth]{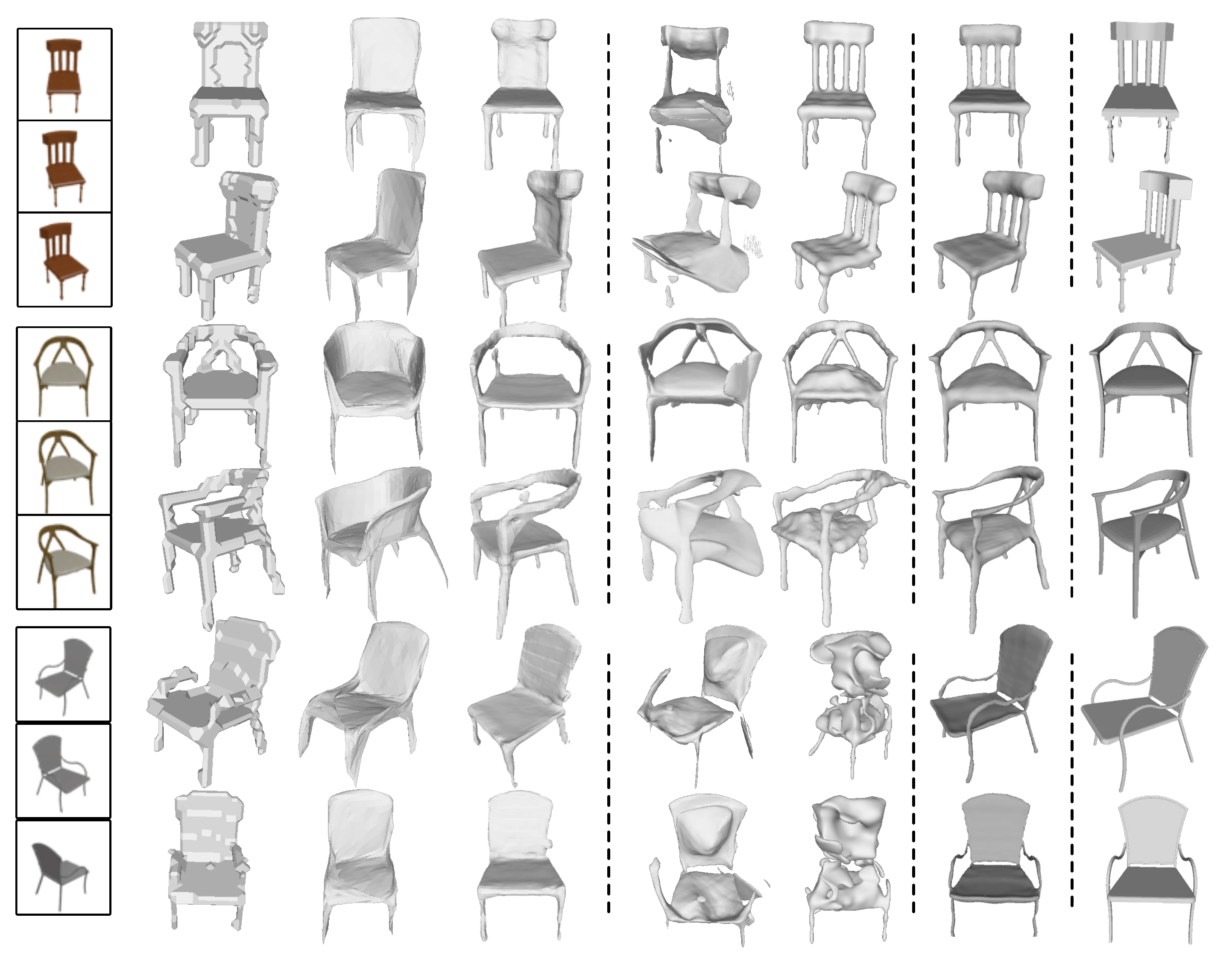}\vspace{-0.5cm}\\
 
    \vspace{-0.1cm}
 
    \caption{
    Qualitative comparisons among sparse multi-view reconstruction results generated by different methods.
    For each instance, the leftmost three images are the only given images, and the rightmost is the ground truth, dubbed as ``GT''.
    We show two different views for each generated object surface, including one from the view of the first image and one completely different view from the three images.
    We use dashed lines to distinguish the methods based on different concepts.}
    \label{fig:mvs_comparing}
     \vspace{-0.1cm}
\end{figure*}

\vspace{0.1cm}
\noindent\textbf{Comparisons.}
We compare our proposed method with the state-of-the-art approachs including LSM \cite{learning-lsm}, P2M++ \cite{wen2019pixel2meshplus}, DISN \cite{NIPS2019_8340}, DVR \cite{DVR} and IDR \cite{yariv2020multiview}.
Specifically, LSM, P2M++, and DISN rely on retrieving the most likely shape priors via forward pass and are the representative works of voxel-, mesh- and implicit function-based approaches respectively.
The other two methods can be considered as optimization-based methods here, which optimize a specific shape using physical constraints.
IDR \cite{yariv2020multiview} achieves the state-of-the-arts for multi-view reconstruction by jointly learning geometry, camera parameters and a differentiable renderer based on multi-view consistency.
DVR \cite{DVR} proposes a differentiable renderer for implicit field that enables unsupervised learning of 3D shape with 2D-to-3D consistency.
In particular, our approach employs the differentiable renderer from DVR in our network training.
We randomly select $50$ instances with $3$ views for each object from the testing set per category, and perform sparse multi-view reconstruction.
Results of these methods are obtained either by using their released codes (if available) or reproducing their methods (multi-view setting in DISN). In both cases, we report their best performance during hyper-parameter tuning.

In Table \ref{table:mvs_comparing}, we compare different methods under the metrics of Chamfer and F-score.
Our proposed method significantly outperforms all the alternative methods in both metrics.
We demonstrate the visual comparison results in Figure~\ref{fig:mvs_comparing}, where we show the reconstructed shapes of several randomly selected instances.
As seen from the results, our method infers the most accurate 3D shape given only very sparse images.
The retrieval based approaches, including LSM, P2M++, and DISN, can recover the rough shape and structure but struggle to capture fine-scale geometry details.
This is primarily due to that they heavily rely on the pre-trained prior and have difficulty generalizing to the unseen data.
The results from DVR \cite{DVR} and IDR \cite{yariv2020multiview}  in some cases perform well from the same views of input images (as shown in the first row of each instance), but appear significantly worse in the views without supervision.
This is because that the training of DVR and IDR only resort to the physical constraints, such as multi-view consistency or 2D-to-3D correspondence. 
As a result, their networks are prioritized to memorize the image-to-shape correspondence but lacks of learning a strong shape prior.
This leads to corrupted results of DVR and IDR as shown in the third group of Figure~\ref{fig:mvs_comparing}.
Since our approach leverages both the pre-trained shape prior and the novel optimization scheme, we are able to precisely reconstruct the intricate and thin structures, e.g, the thin chair legs and back structure, while ensuring a plausible shape.
For more qualitative results, please refer to the appendix.
\begin{table}
\begin{center}
\begin{tabular}{|l|c|c|c|c|}
\hline
\multirow{2}{*}{Methods} & \multicolumn{2}{c|}{CD, mean $(10^{-3})$} &  \multicolumn{2}{c|}{F-score}\\
\cline{2-3} \cline{4-5}
~ & Chair & Lamp & Chair & Lamp \\
\hline\hline
LSM \cite{learning-lsm} &7.36 &6.32 & 27.43& 25.89\\
P2M++ \cite{wen2019pixel2meshplus} &8.41 & 7.89& 37.23& 32.15\\
DISN \cite{NIPS2019_8340} &2.75 & 15.29 & 52.47& 26.03\\
\hline
Ours & $\bm{1.75}$ & $\bm{5.44}$&$\bm{62.34}$ &$\bm{36.20}$\\
\hline
\end{tabular}
\end{center}
\vspace{-0.3cm}
\caption{
Comparative results of different methods for sparse view reconstruction.
Performance is measured in terms Chamfer and F-score over $50$ instances.
$10^{-3}$ refers to the magnitude.
We only show the comparative results with learning-based methods because optimization-based methods are unstable in some instances, which leads to terribly bad numerical results.
}
\label{table:mvs_comparing}
\vspace{-0.4cm}
\end{table}

\begin{figure}
    \begin{tabular}{p{27pt}p{27pt}p{27pt}p{27pt}p{27pt}p{27pt}}
        DISN & Ours & \ GT & \ IDR & Ours & \ GT
    \end{tabular}
    \subcaptionbox{Single View}
    {\includegraphics[width=0.23\textwidth]{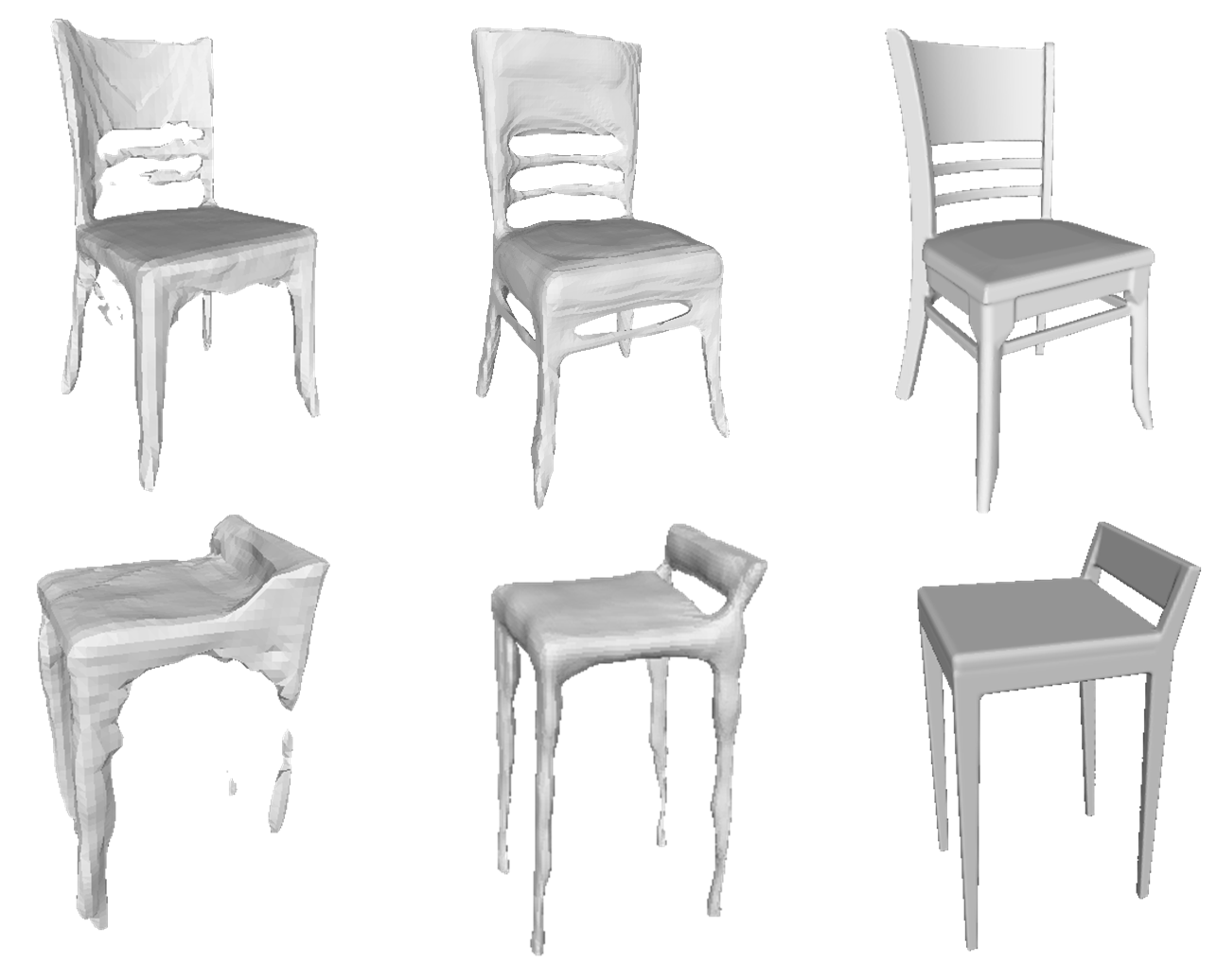}}
    \subcaptionbox{Dense View}
    {\includegraphics[width=0.23\textwidth]{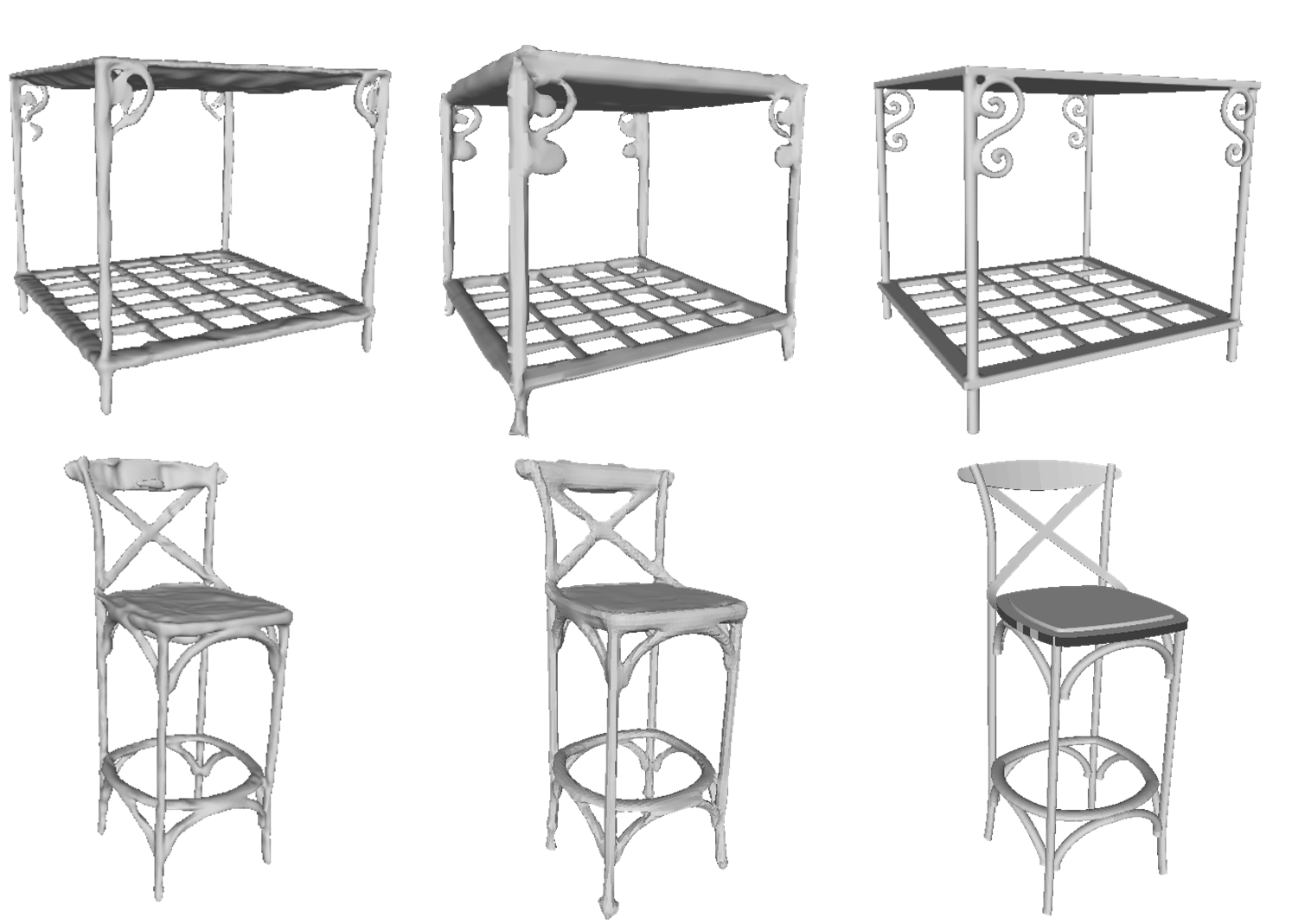}}
    \centering
    \caption{
        Qualitative comparisons among reconstructed results for the extreme cases of single view and dense views.
        We compare with the state-of-the-art method learning-based DISN \cite{NIPS2019_8340} under single view setting while comparing with the state-of-the-art optimization-based method  IDR\cite{yariv2020multiview} under dense view setting.
    }
    \label{fig:diff_views_com}
    \vspace{-0.4cm}
\end{figure}

\vspace{0.1cm}
\noindent\textbf{Control Studies on Number of Views.}
 \begin{figure}
    \begin{tabular}{p{37pt}p{37pt}p{37pt}p{37pt}p{37pt}}
    \ \ $1$ view & \ $2$ views & \ $6$ views & $12$ views\ \ & \ \ \ GT \\
    \end{tabular}
    \centering
    \includegraphics[width=0.49\textwidth]{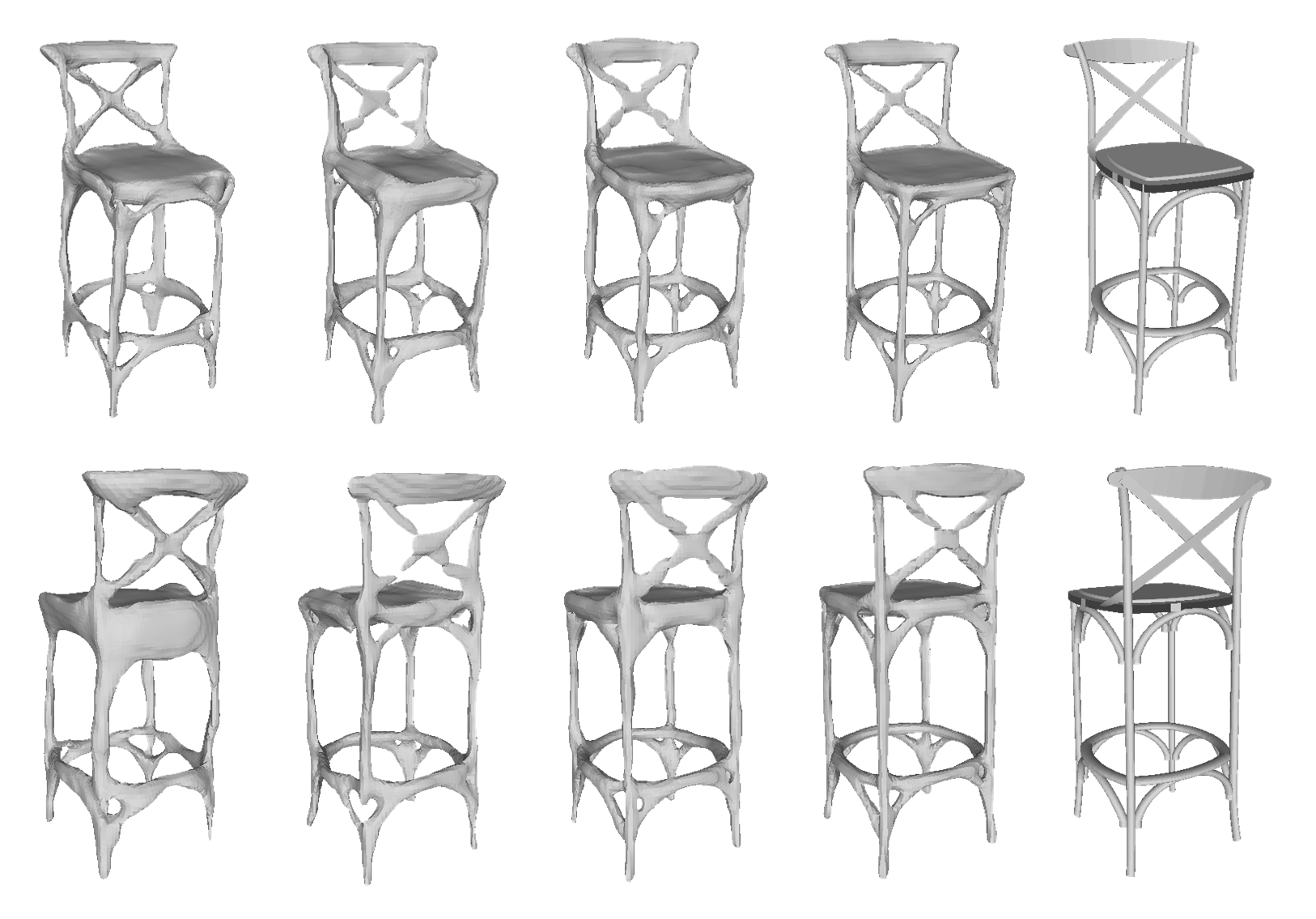}\vspace{-0.3cm}\\
    \caption{
        Example results of control studies on different number of views.
      }
    \label{fig:diff_views}
    \vspace{-0.1cm}
\end{figure}
We also conduct ablation study to evaluate the performance of our approach given different number of input views.
In particular, we test our approach using $1$, $2$, $6$ and $12$ views.
As can be seen in Figure \ref{fig:diff_views}, our approach can produce robust reconstruction with only a single view.
In addition, with more views available, the quality of our reconstruction can be further improved.
When 12 views are present, we can achieve similar quality of reconstruction with that of the stereo-based approach.
It indicates that the learning-based approach can benefit a lot optimizing the pre-trained prior according to the physical constraints,  
In Figure \ref{fig:diff_views_com}, we further compare our approach with DISN, which is specialized for singe-view reconstruction, and with IDR that excels at using dense multi-views.
In comparisons, our approach can achieve similar or even better reconstructions.

\subsection{Point Cloud Reconstruction} 
\label{sec:completion}

\begin{figure}
    \begin{tabular}{p{40pt}p{35pt}p{35pt}p{35pt}p{35pt}}
        Point set & IFNet\cite{chibane2020implicit} & IGR\cite{gropp2020implicit} &  Ours & GT
    \end{tabular}
    \centering
    \includegraphics[width=0.49\textwidth]{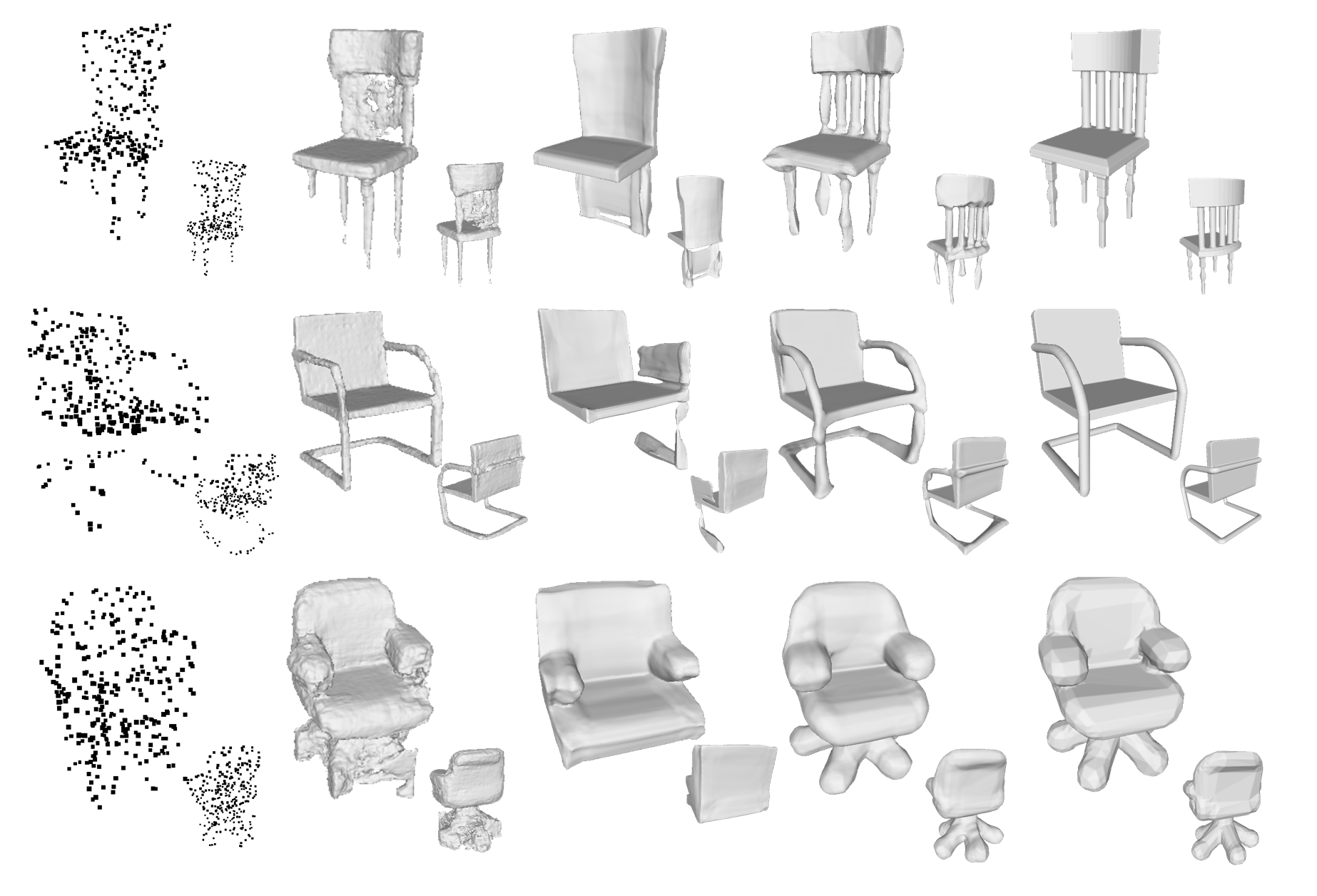}
    \vspace{-0.7cm}
    \caption{
    Qualitative comparisons between reconstruction results generated by different methods.
    The reconstruction results are based on the leftmost sparse point cloud. }
    \label{fig:pc_comparing}
    \vspace{-0.2cm}
\end{figure}

We compare our approach with the state-of-the-art point reconstruction approaches: IFNet \cite{chibane2020implicit} and IGR \cite{gropp2020implicit}.
 We test all the approaches using a highly sparse point cloud consisting of 300 points.
The qualitative and quantitative results can be found in Figure \ref{fig:pc_comparing} and Table \ref{table:pc_comparing} respectively.
Compared to IFNet and IGR, our approach can better reconstruct the intricate geometry details, such as the thin slats in the chair back (1st row), with quality close to the ground truth.
In contrast, IGR fails to generate the thin chair legs (2nd row) while IFNet suffers from artifacts (1st and 3rd rows).
 In particular, we achieve these results by first searching for the instance shape code that best corresponds to the given point cloud, and then gradually updates the parameters of the implicit prior for the shape fitting.
The initial latent code plays an important role for regularizing the subsequent optimization and provide an accurate initialization to optimize from.

\begin{table}
\begin{center}
\begin{tabular}{|l|c|c|c|c|}
\hline
\multirow{2}{*}{Methods} & \multicolumn{2}{c|}{CD, mean $(10^{-3})$} &\multicolumn{2}{c|}{Normal-Consis.}\\ 
\cline{2-3} \cline{4-5}
~ & Chair & Car & Chair & Car\\
\hline\hline
IFNet \cite{chibane2020implicit} & 0.72 &  1.68  & 0.84 & 0.85 \\
IGR \cite{gropp2020implicit} &1.55 & 0.72  &0.75 & 0.89 \\
\hline
Ours & $\bm{0.64}$ & $\bm{0.28}$ & $\bm{0.86}$& $\bm{0.90}$  \\
\hline
\end{tabular}
\end{center}
\vspace{-0.3cm}
\caption{
Comparative results of different methods for shape reconstruction from sparse point clouds.
Performance is measured in terms of Chamfer distance and normal consistency over $50$ instances.
$10^{-3}$ refers to the magnitude.
}
\label{table:pc_comparing}
\vspace{-0.4cm}
\end{table}

\subsection{Ablation Studies}
\begin{figure}
    \begin{tabular}{c c c c}
    Image & Our method & \makecell{W/O\\pretraining $\theta$} & \makecell{W/O\\optimizing $\theta$}
    \end{tabular}\vspace{-0.2cm}\\
    \centering
    \includegraphics[width=0.48\textwidth]{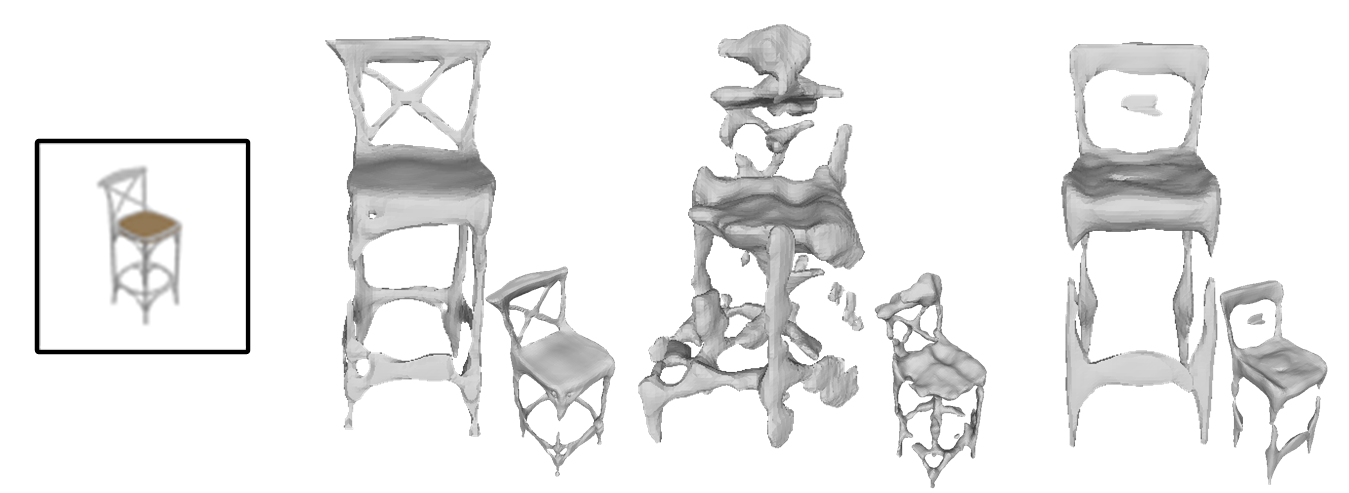}\vspace{-0.2cm}\\
    \caption{
    Example results of ablation study among our proposed method, optimizing both the latent code and the randomly initialized network, and optimizing the latent code only. }
    \label{fig:ablation}
    \vspace{-0.4cm}
\end{figure}

 In this section, we perform ablation studies to evaluate the efficacy of our proposed pipeline.
All the following experiments are conducted in the context of multi-view reconstruction.
Results are shown in Figure \ref{fig:ablation}, including our proposed method (optimizing both the latent code and the parameters of the pre-trained network), optimizing from random initialized network and optimizing only latent code.
The reconstructions are based on the leftmost image (single view reconstruction), and then in turns our proposed method, the one without pre-trained network parameters and the one without optimizing network parameters.
We show two different views for each generated object surface, including one from the view of the image and one from a completely different view.
\vspace{0.1cm}

\noindent\textbf{Optimize latent code only.}
\vspace{0.1cm}
One of the keys to our approach is optimizing both the latent code and the parameters of the network, making them adapt to the given observation.
Though DeepSDF~\cite{park2019deepsdf} has shown promising results by only optimizing latent code,
we find in many cases optimizing latent code, such as the 4th column in Figure~\ref{fig:ablation}, fails to faithfully reconstruct the geometric details, especially for unseen data.
In contrast, our approach can achieve better results by jointly optimizing the latent code and the shape prior.

\noindent\textbf{Optimizing from randomly initialized parameters.}
The other key for our approach is to pre-train the parameters of the network via an additional dataset to obtain a good initialization.
As seen in Figure~\ref{fig:ablation}, optimizing a randomly initialized network (the 3rd column) fails to generate a plausible shape.
Without the proposed pre-training, the subsequent optimization may deviate from a plausible searching path especially for highly ill-posed problems, such as single-view 3D reconstruction.

	\section{Conclusions and Discussions}
We have presented a new learning framework for 3D modeling and reconstruction that profits from both the advantages of learning-based and optimization-based approaches.
We have shown that by jointly optimizing the pre-trained prior and the latent code at test time, according to the physical constraints, is a promising avenue to greatly improve the generality of a deep prior.
To ensure the optimization would lead to reasonable result, we proposed that a simple $l_2$ regularization mechanism plays an important role in regularizing the searching space.
Our experiments and evaluations have shown that our approach can generalize significantly better to unseen data compared to alternative approaches, especially in presence of sparse or highly collapsed inputs.
Despite these promising directions, our method is currently more expensive that alternatives. It would be an interesting avenue to accelerate the optimization with additional hypernetworks.
In addition, we are still lacking a theoretical analysis of the working principle of our approach, which will be our next focus.


	
	{\small
		\bibliographystyle{ieee_fullname}
		\bibliography{paper}
	}
	
	\clearpage
	\appendix
  \noindent  \textbf{\Large Appendix}
  \vspace{0.5cm}
  
This appendix provides experimental details and more qualitative experimental results that are supplementary to the main paper.
We first describe the network details of different applications (Section \ref{app:component_mvs} and Section \ref{app:component_sdf_pc}).
We then introduce how we perform pre-training on the auxiliary dataset (Section \ref{app:method_training}).
Further, we provide experimental details (Section \ref{app:implementation}).
Finally, more ablation study for $l_2$ norm constraint are presented (Section \ref{app:ablation}), and additional qualitative results are presented (Section \ref{app:quality}).

	\section{Components Details for Sparse Multi-view Reconstruction} \label{app:component_mvs}
The overall pipeline is depicted in Figure \ref{fig:sup_overview}.
There are five main components in our proposed sparse multi-view reconstruction method, including an implicit shape network representing the  implicit field of the 3D geometry, a neural texture network predicting the implicit texture field of the object, a hypernetwork learning the categorical hyper-parameters, a latent code space distinguishing different instances among the same category and a differentiable renderer converting the 3D shape into 2D images.
In particular, we aim to minimize the difference between our synthetic 2D images/silhouette and the given images/silhouette, by optimizing the parameters of the hypernetwork and the latent code (through the implicit shape network and the neural texture network).

\subsection{Implicit Shape Network}
A 3D geometrical shape $\mathcal{O}$ can implicitly defined as the zero level set of a neural network $f_{\text{\tiny{geo}}}:\mathbb{R}^3\to\mathbb{R}$, as
\begin{equation}
\mathcal{O} = \{\bm{p} \in \mathbb{R}^3 \vert f_{\text{\tiny{geo}}}(\bm{p};\bm{\phi_1})=0\},
\end{equation}
where $\bm{\phi}_1$ denotes the parameters for the neural network, $\bm{p}$ denotes the coordinate of a point in the 3D space.
Note that $f_{\text{\tiny{geo}}}(\cdot)$ in our pipeline works as a generator to represent the shapes, \ie, we could generate various shapes by controlling $\bm{\phi}_1$.
With differentiable renderers, the implicit field of shape itself can be rendered as the silhouette of the object, which enables 2D supervision by comparing the rendered outcome with the ground-truth 2D mask.

\subsection{Implicit Texture Network}

\begin{figure}
	\centering
	\includegraphics[width=.49\textwidth]{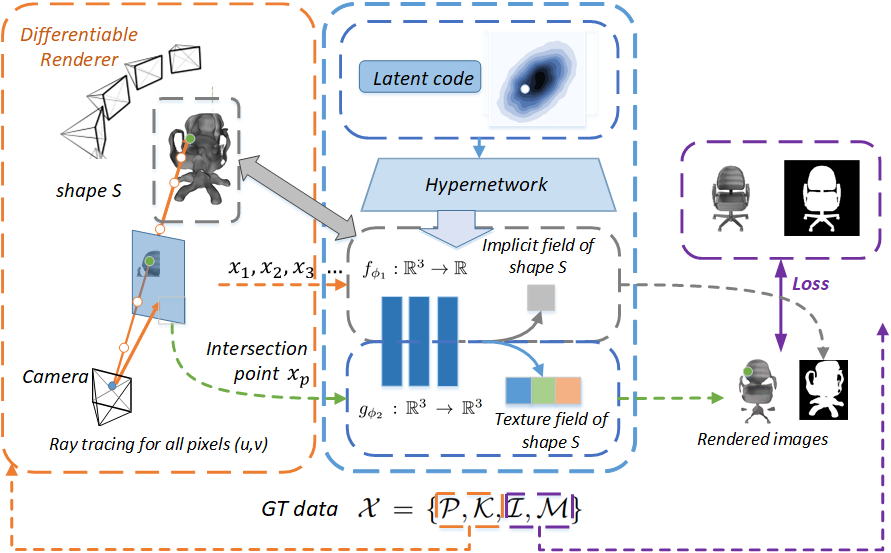}
	\caption{
	An overview of our proposed reconstruction pipeline of sparse images reconstruction.
	}
	\label{fig:sup_overview}
\end{figure}

We further use a neural network $g_{\text{tex}}:\mathbb{R}^3\to\mathbb{R}^3$ to regress the texture field \cite{oechsle2019texture} of the object, whose input is a point in the 3D space, and output is the RGB color of this point.
Formally, $g_{\text{tex}}$ can be defined as,
\begin{equation}
\bm{c} = g_{\text{tex}}(\bm{p};\bm{\phi_2}),
\end{equation}
where $\bm{\phi}_2$ denotes the parameters for the neural network, and $\bm{c}$ denotes the color of the point $\bm{p}$.
Note that we follow \cite{DVR} to let $g_{\text{tex}}(\cdot)$ share the feature extractor with the implicit shape network $f_{\text{\tiny{geo}}}(\cdot)$, \ie, they share a majority of parameters, as depicted in Figure \ref{fig:sup_overview}.
Hence, we are able to connect the geometric shape of an object with its color, which allows us to further use color supervision to reconstruct the shape.

\subsection{Hypernetwork}
Hypernetwork is a neural network that regresses the parameters of another neural network.
In this work, we adopt a hypernetwork to provide semantic control of per-category shapes, whose child networks are the neural networks of $f_{\text{\tiny{geo}}}$ and $g_{\text{\tiny{tex}}}$ as mentioned above.
Formally, a hypernetwork $h_{\text{\tiny{hyper}}}:\mathbb{R}^{|\bm{z}|}\to\mathbb{R}^{|\bm{\phi}|}$ can be defined as
\begin{equation}
\bm{\phi} = h_{\text{\tiny{hyper}}}(\bm{z};\bm{\theta}),
\end{equation}
where $\bm{\theta}$ denotes the parameters of the hypernetwork and $\bm{z}$ denotes the latent shape code specifically for an instance.
And $\bm{\phi} = \{\bm{\phi_1} \cup \bm{\phi_2}$\} denotes the union of the parameters of $f_{\text{\tiny{geo}}}(\cdot)$ and $g_{\text{\tiny{tex}}}(\cdot)$.
$\bm{\theta}$ here can be regarded as the prior of certain category learned from the training data $\{\mathcal{X}^j\}_{j=1}^M$.
Ideally, given arbitrary shape code $\bm{z}$, $h_{\text{\tiny{hyper}}}(\cdot)$ would output the parameters for $f_{\text{\tiny{geo}}}(\cdot),g_{\text{\tiny{tex}}}(\cdot)$ that represents reasonable shapes in the given category.

\subsection{Latent Shape Space}
The latent shape space is constructed via the auto-decoder \cite{park2019deepsdf} of hypernetwork.
Individual latent code in the latent space stands for a specific shape of its corresponding instance.
In other words, given observed images, we assume that there exists a shape code corresponding to the underlying 3D surface represented by these images.
We encourage the prior distribution over $p(\bm{z})$ to be a zero-mean multivariate-Gaussian with a spherical covariance $\sigma^2 I$ during training to construct a compact and meaningful shape manifold.

\subsection{Differentiable Renderer}\label{app:component_mvs_DR}
For differentiable renderer, we mainly leverage the work of Niemeyer \etal \cite{DVR}. 
With \cite{DVR, yariv2020multiview}, we could get the 2D silhouette and 2D images of the object.
To be more specific, given camera extrinsic and intrinsic, a ray will be cast from a pixel in the synthetic 2D image towards the 3D object surface to find the intersected point.
The color of the intersected point is regarded as the color of this pixel.
Collectively, the synthetic 2D images can be generated.

 	\section{Components Details for Shape Auto-encoding and Shape Completion} \label{app:component_sdf_pc}

Comparing to the sparse multi-view reconstruction, shape reconstruction and point cloud reconstruction share a much more straightforward pipeline.
Three of the five components mentioned above are retained, including the implicit neural network, the hypernetwork and the latent shape-code space.
We minimize the difference between our generated signed distance field / point cloud and the given observations according to corresponding losses.
 
 	\section{Training on Auxiliary Dataset}\label{app:method_training}
In this section, we will introduce the pretraining strategy we adopted for different applications.
Let $\bm{\theta}_0$ denote the parameters of $\bm{\theta}$ learned from a collection of $\{\mathcal{X}^j\}_{j=1}^{N}$ (training set) representing $N$ different object surfaces of the same category.

\subsection{Pre-training for Shape Auto-Encoding}
In the training phase of this task, given a collection of $\{\mathcal{X}^j\}_{j=1}^{N}$ representing $N$ different signed distance fields of the same category, we drop the $l_2$ norm constrain on $\bm{\theta}$ in Eq. (\ref{eq:regu_mvs}) directly without other physical constraints, leaving only the constraint on $\bm{z}$.
The overall objective function can be formulated as 
\begin{equation}\label{eq:overall_training_sdf}
    \bm{\theta}_0 = \arg\min_{\bm{\theta}} \sum_{j\in\{1,\ldots,N\}} E(f_{\bm{\theta}}(\bm{z}^{j}); \mathcal{X}^j) + \lambda_T\cdot\frac{1}{\sigma^2}\|\bm{z}^{j}\|_2,
\end{equation}
where $\lambda_{T}$ denotes the weighted parameter.

\subsection{Pre-training for Multi-view Reconstruction }
For multi-view reconstruction, denser views are provided for each $\mathcal{X}$ in the training phase than in the inference phase to make the learning more feasible.
We mainly follow Eq. (\ref{eq:overall_mvs}) to learn the categorical prior for $\theta$.
The only difference is that we replace the $l_2$ norm constrain on $\bm{\theta}$ in Eq. (\ref{eq:regu_mvs}) with a physical constraints of smoothness to encourage the normals within a surface local patch pointing to the similar directions, which can be formally defined as 
\begin{equation}\label{eq:regu_training_mvs}
    R_T(f_{\bm{\theta}}(\bm{z})) = \frac{1}{\sigma^2}\|\bm{z}\|_2 + \lambda_{T_{\hat{\bm{n}}}}\cdot\sum_{\bm{p}_i\in\hat{\mathcal{O}}}( \hat{\bm{n}}(\bm{p}_i) -  \frac{1}{k}\sum_{\bm{p}\in\mathcal{N}_{\bm{p}_i}} \hat{\bm{n}}(\bm{p})),
\end{equation}
where $\lambda_{T_{\hat{\bm{n}}}}$ is the weighted parameter.
$\hat{\bm{n}}(\bm{p})$ is the estimated normal for point $\bm{p}$ sampled from the estimated object surface $\hat{\mathcal{O}}$ following \cite{DVR}, and $\mathcal{N}_{\bm{p}_i}$ are the set of $k$ nearest neighboring points of point $\bm{p}_i$.
And the overall objective function can be formulated as 
\begin{equation}\label{eq:overall_training_mvs}
    \bm{\theta}_0 = \arg\min_{\bm{\theta}} \sum_{j\in\{1,\ldots,N\}} E(f_{\bm{\theta}}(\bm{z}^{j}); \mathcal{X}^j) + \lambda_T\cdot R_T(f_{\bm{\theta}}(\bm{z}^{j})) ,
\end{equation}
where $\lambda_T$ denotes the weighted parameter and $E(f_{\bm{\theta}}(\bm{z}^{j}); \mathcal{X}^j)$ is the same as in Eq. (\ref{eq:energy_mvs}).

\subsection{Pre-training for Shape Completion}
The regularizer for point cloud shape completion can be formulated as 
\begin{equation}\label{eq:regu_training_pc}
    R_T(f_{\bm{\theta}}(\bm{z})) = \frac{1}{\sigma^2}\|\bm{z}\|_2 +  \lambda_{T_g}\cdot\mathbb{E}_{\bm{p}}(\|\nabla_{\bm{p}}f(\bm{z},\bm{p}_i)\|_2-1)^2,
\end{equation}
where $\lambda_{T_g}$ is the weighted parameter.
And the overall objective function can be formulated as 
\begin{equation}\label{eq:overall_training_pc}
    \bm{\theta}_0 = \arg\min_{\bm{\theta}} \sum_{j\in\{1,\ldots,N\}} E(f_{\bm{\theta}}(\bm{z}^{j}); \mathcal{X}^j) + \lambda_T\cdot R_T(f_{\bm{\theta}}(\bm{z})),
\end{equation}
where $\lambda_{T}$ is also the weighted parameter.
 	\section{Implementation Details}\label{app:implementation}
As mentioned in Section \ref{app:component_mvs} and Section \ref{app:component_sdf_pc}, for all the applications, we adopt the architecture of auto-decoder.
For all experiments in main paper we used a network composed of $9$ fully connected layers with ReLU activation for the implicit shape field (texture field). 
The network takes a 3D point as input, with a skip connection included at the $4$ layer for the input point, and outputs occupancy value or signed distance value. 
And the weight matrices of the implicit network are generated by the hypernetwork, which takes in the latent vector of $256$ dimensions and outputs weights.
The hypernetwork is parameterized as a two-layer perceptron ($256$ units per layer) with ReLU activation and layer normalization \cite{ba2016layer} before each non-linearity.  
Given a pre-trained model and the input observations, we firstly estimate the optimal latent code with the corresponding objective functions, and then jointly optimize the parameters in the hypernetwork and the latent code according to the same objective functions.
For all experiments, we adopt Adam \cite{kingma2014adam} with $\beta_1=0.9$, $\beta_2=0.999$ to optimize the parameters in the hypernetwork with respect to the objective function.
We decay the learning rate at $1k$, $1.5k$ and $2k$, with totally $3k$ epochs, where the initial learning rate is $0.0001$ for sparse multi-view reconstruction; with fixed learning schedule of $0.001$ and totally $1k$ epochs for shape auto-encoding; with fixed learning schedule of $0.005$ and totally $1.5k$ epochs for shape completion.
And note that for better results, we also adopt the strategy of early stop \cite{yao2007early}.
We use the default values of $\lambda_c=0.5$ in Eq. (\ref{eq:energy_mvs}) and $\lambda_{\theta}=0.1$ in Eq. (\ref{eq:regu_mvs}); $\lambda_{T_{\hat{\bm{n}}}}=2.0$ in Eq. (\ref{eq:regu_training_mvs}).
We use $\lambda_{g}=1.0$ in Eq. (\ref{eq:regu_pc}) , and $\lambda_{T_g}=1.0$ in Eq. (\ref{eq:regu_training_pc}). For all experiments, we use $\lambda=0.1$, $\lambda_T=1.0$.
$3$ views are provided in multi-view reconstruction during inference without further illustration, and $24$ views are provided for training; $300$ points are provided in shape completion during inference without further illustration, and $150k$ points are provided for training; and we adopt a discrete signed distance field of $250k$ signed distance values for shape auto-encoding for both training and inference.
Besides, for the task of multi-view reconstruction,  we use the non-linear maps of inputs (positional encoding) \cite{mildenhall2020nerf} to improve the learning of high-frequencies.
And we converted the implicit field model into a mesh by using Marching Cubes \cite{lorensen1987marching} with $128^3$ resolution.

 	\section{Ablation Study on $l_2$ Constraint}\label{app:ablation}

Additional ablation study of $l_2$ constraint on the network parameters can be found in this section.
For better consistency, the following experiments is also conducted in the context of multi-view reconstruction.

\begin{figure}[h]
    \begin{tabular}{p{40pt}p{55pt}p{55pt}p{55pt}}
    Image & W/O $l_2$ & With $l_2$ & GT
    \end{tabular}\vspace{-0.1cm}\\
    \centering
    \includegraphics[width=0.49\textwidth]{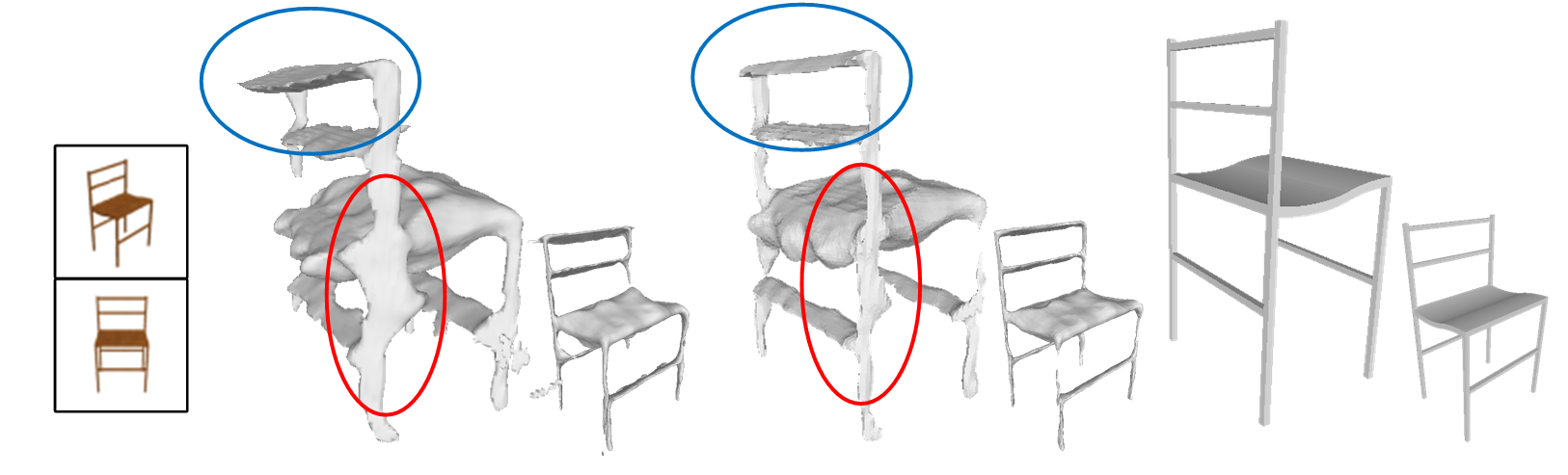}\vspace{-0.3cm}\\
    \caption{
    Example results of ablation study among our proposed method. }
    \label{fig:ablation_l2}  
    \vspace{-0.3cm}
\end{figure}

With the technique of differentiable renderer, we optimize the neural implicit field representing the object surface according to the given sparse observations until the rendered images are exactly alike the given images.
And as can be seen in Figure \ref{fig:ablation_l2}, without using $l_2$, the results might overfit to the views with given observations, while causing error from the other views without supervision.
Alone with the prior, we can hardly expect the optimization to be perfectly proper, since the observations are very sparse, continuous fitting might will eventually destroy the initial prior.
Thus, the $l_2$ constraint is simple but effective method avoiding such overfitting.

 	\section{More Qualitative Results}\label{app:quality}

\begin{figure*}
    \centering
    \begin{tabular}{
    p{50pt}p{50pt}p{50pt}p{50pt}p{60pt}p{50pt}p{50pt}p{50pt}
    }
   \ \  \  Images & LSM\cite{learning-lsm} & P2M++\cite{wen2019pixel2meshplus} & DISN\cite{NIPS2019_8340} & \ \  DVR\cite{DVR} &  IDR\cite{yariv2020multiview} &  Ours & \ \     GT 
    \end{tabular}\vspace{-0.1cm}\\
    \includegraphics[width=.99\textwidth]{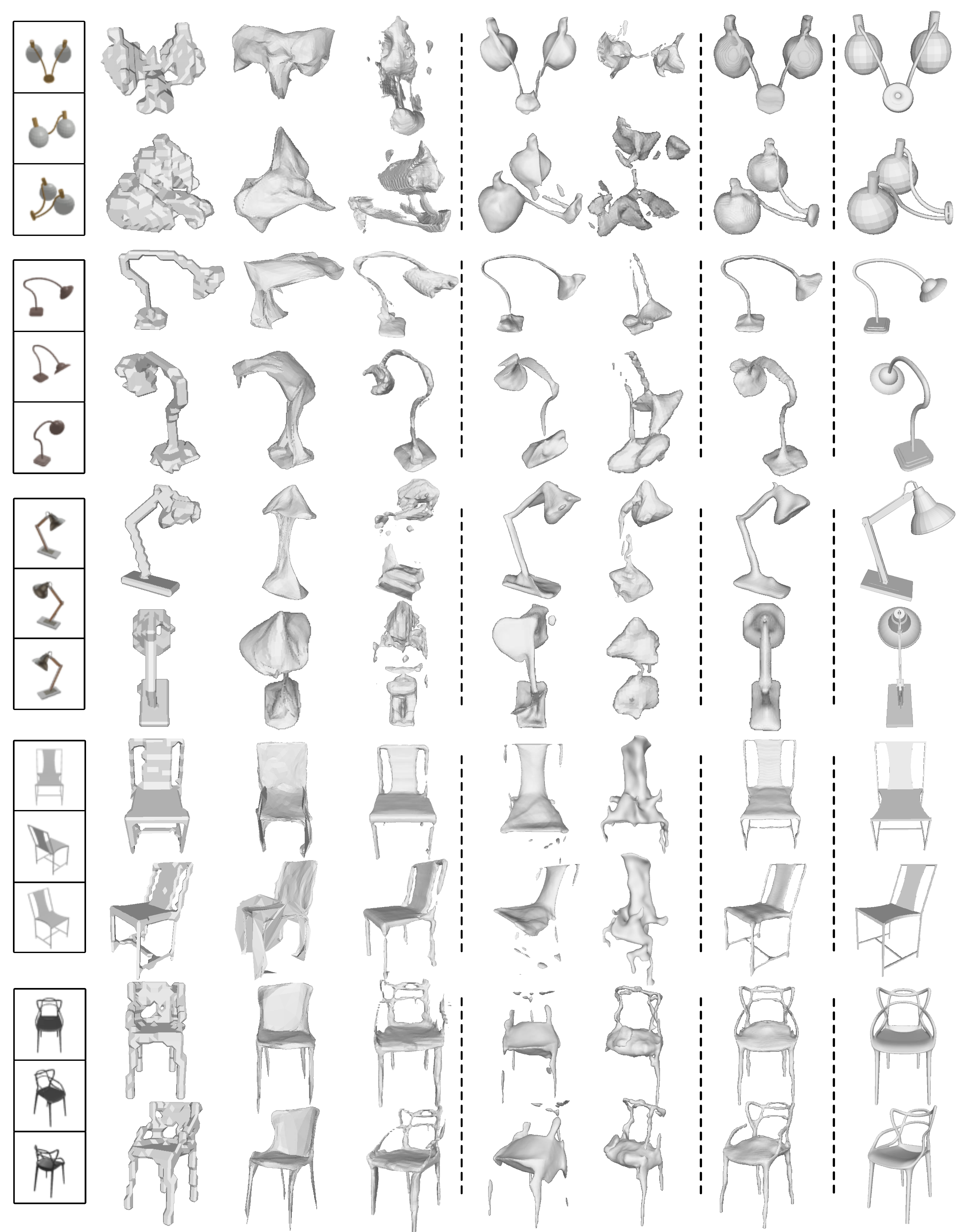}\vspace{-0.2cm}\\
  \caption{
    More qualitative comparative results for sparse view reconstruction generated by different methods.
    }
    \label{fig:mvs_comparing_supp}
    \vspace{-0.2cm}
\end{figure*}

\begin{figure*}[h]
    \begin{minipage}[c]{0.08\textwidth}
        \raggedright
        DeepSDF\cite{park2019deepsdf}
    \end{minipage}
    \begin{minipage}[c]{0.50\textwidth}
        \raggedright 
        \includegraphics[height=0.31\textwidth]{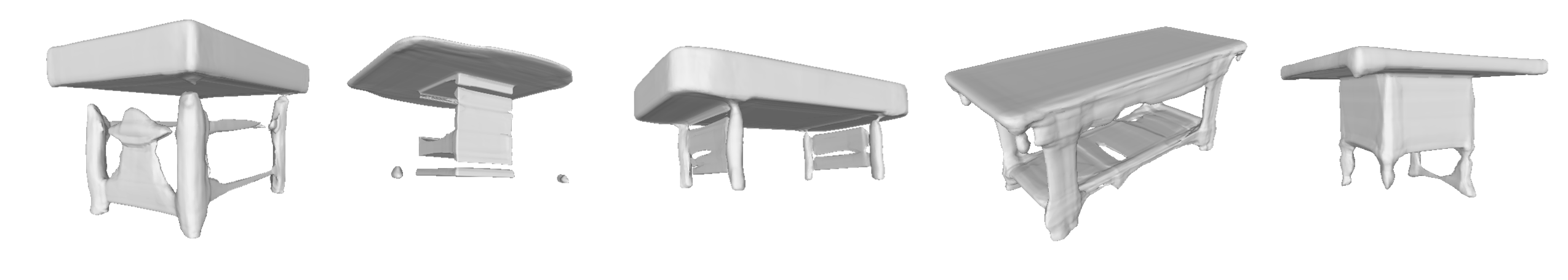}
    \end{minipage}\vspace{-0.3cm}\\
    \begin{minipage}[c]{0.08\textwidth}
        \raggedright
        Ours
    \end{minipage}
    \begin{minipage}[c]{0.50\textwidth}
        \raggedright 
        \includegraphics[height=0.31\textwidth]{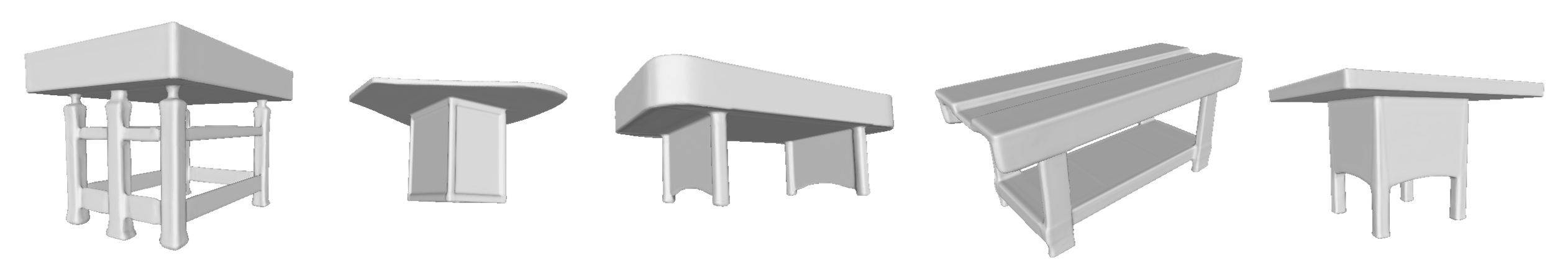}
    \end{minipage}\vspace{-0.3cm}\\
    \begin{minipage}[c]{0.08\textwidth}
        \raggedright
        GT
    \end{minipage}
    \begin{minipage}[c]{0.50\textwidth}
        \raggedright 
        \includegraphics[height=0.31\textwidth]{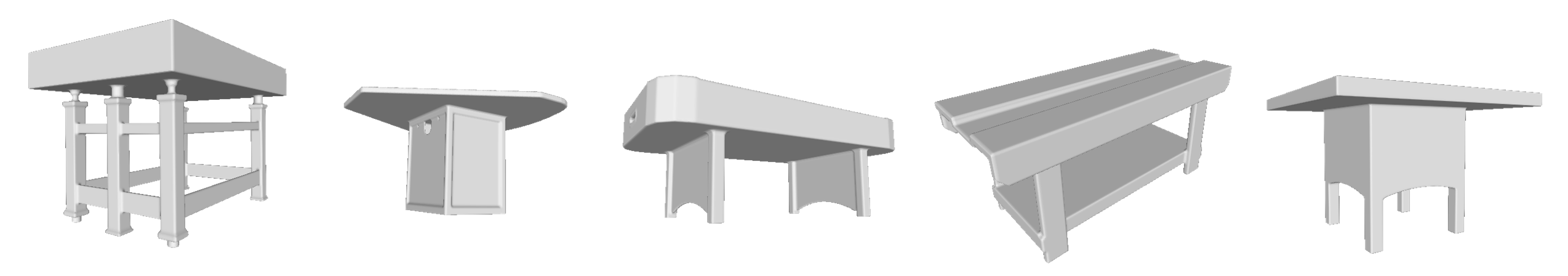}
    \end{minipage}\vspace{-0.3cm}\\
     \caption{
   More qualitative comparative results for shape auto-encoding generated by different methods.
 }
    \label{fig:sdf_comparing_supp}
    \vspace{-0.2cm}
\end{figure*}

\begin{figure*}
    \centering
    \begin{tabular}{
 p{70pt}p{70pt}p{70pt}p{70pt}p{90pt}
    }
    Point set & \ \ \ \ IFNet\cite{chibane2020implicit} & \qquad \ \ IGR\cite{gropp2020implicit} & \qquad\quad\ \ \ \ Ours & \qquad\qquad\qquad GT
    \end{tabular}
    \centering
    \includegraphics[width=0.99\textwidth]{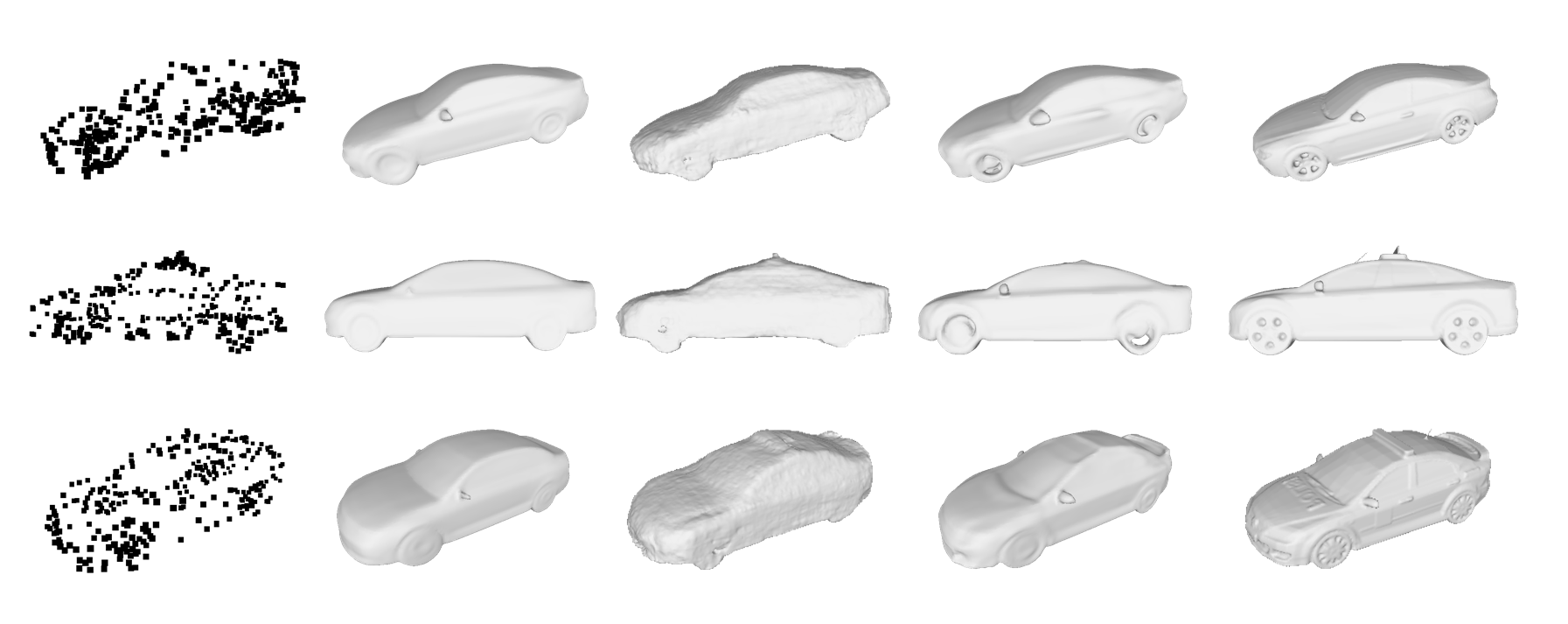}
    \vspace{-0.7cm}
    \caption{
  More qualitative comparative results for sparse point cloud completion generated by different methods.
    }
    \label{fig:pc_comparing_supp}
    \vspace{-0.3cm}
\end{figure*}

In this section, we provide additional qualitative experimental results.
The reconstruction results of more categories for multi-view reconstruction are provided in Figure \ref{fig:mvs_comparing_supp}.
The results of shape auto-encoding and shape completion are provided in Figure \ref{fig:sdf_comparing_supp} and Figure \ref{fig:pc_comparing_supp} respectively.

	\end{document}